\PassOptionsToPackage{table}{xcolor}
\documentclass[sigconf]{aamas}
\makeatletter
\def\aamas@legacy@editors{\fi}
\ifx\acmConference@editors\aamas@legacy@editors
  \expandafter\@firstofone
\else
  \expandafter\@gobble
\fi
{\acmConference@editors\gdef\acmConference@editors{}}
\let\aamas@legacy@editors\@undefined
\makeatother

\usepackage[utf8]{inputenc}
\usepackage{balance}
\usepackage{enumitem}
\usepackage{xurl}
\usepackage{array}
\usepackage{booktabs}
\usepackage[table]{xcolor}
\usepackage{tabularx}

\newcommand{\xmark}{\ding{55}}
\setlist{nosep,leftmargin=*}

\AtEndPreamble{%
  \hypersetup{
    pdfauthor={},
    colorlinks=true,
    linkcolor=blue,
    citecolor=black,
    urlcolor=blue,
    pdfborder={0 0 0},
    pdfborderstyle={}
  }%
}

\newsavebox{\ManuscriptLinkBox}
\DeclareRobustCommand{\ManuscriptLinkFrame}[2]{%
  \begingroup
    \setlength{\fboxsep}{0.3pt}%
    \setlength{\fboxrule}{0.6pt}%
    \sbox{\ManuscriptLinkBox}{#2}%
    \leavevmode
    \rlap{%
      \kern-\dimexpr\fboxsep+\fboxrule\relax
      \smash{\textcolor{#1}{\fbox{\phantom{\usebox{\ManuscriptLinkBox}}}}}%
    }%
    \usebox{\ManuscriptLinkBox}%
  \endgroup
}
\AddToHook{begindocument/end}{%
  \let\ManuscriptOriginalCiteNumFont\citenumfont
  \renewcommand*{\citenumfont}[1]{%
    \ManuscriptLinkFrame{green}{\ManuscriptOriginalCiteNumFont{#1}}%
  }%
  \let\ManuscriptOriginalEqref\eqref
  \renewcommand*{\eqref}[1]{%
    \begingroup
      \hypersetup{linkcolor=black}%
      \ManuscriptLinkFrame{red}{\ManuscriptOriginalEqref{#1}}%
    \endgroup
  }%
}

\newtheorem{definition}{Definition}
\newtheorem{assumption}{Assumption}

\makeatletter
\gdef\@copyrightpermission{
  \begin{minipage}{0.2\columnwidth}
   \href{https://creativecommons.org/licenses/by/4.0/}{\includegraphics[width=0.90\textwidth]{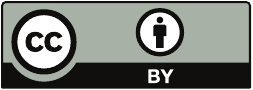}}
  \end{minipage}\hfill
  \begin{minipage}{0.8\columnwidth}
   \href{https://creativecommons.org/licenses/by/4.0/}{This work is licensed under a Creative Commons Attribution International 4.0 License.}
  \end{minipage}
  
}
\makeatother

\setcopyright{ifaamas}
\acmConference[AAMAS 2027]{Proc.\@ of the 26th International Conference
on Autonomous Agents and Multiagent Systems (AAMAS 2027)}{3--7 May 2027}{Hanoi, Vietnam}{M.~Baldoni, F.~Fang, W.~Yeoh, N.~Yorke-Smith (eds.)}
\copyrightyear{2027}
\acmYear{2027}
\acmDOI{}
\acmISBN{}

\acmSubmissionID{}
\submissionType{Research Paper Track}

\title{World Model Science: Self-Organized Criticality, Weak Chaos, and Metastable Belief Dynamics in Long-Horizon LLM Agents}

\author{Xinyuan Song}
\affiliation{
  \institution{Emory University}
  \city{Atlanta}
  \country{USA}
}
\email{xinyuan.song@emory.edu}
\author{Zekun Cai}
\affiliation{
  \institution{The University of Tokyo}
  \city{Tokyo}
  \country{Japan}
}
\email{caizekun@csis.u-tokyo.ac.jp}

\begin{abstract}
Long-horizon LLM agents must maintain and update a world model over extended sequences of observations, actions, decisions and tool calls. We study the dynamics of long horizon agents from three perspectives: self-organized criticality, weak chaos, and metastable belief dynamics. Through extensive experiments on controlled puzzles, tool-use tasks, embodied environments, multi-hop retrieval, general-assistant reasoning, and Game of Life, we study how world-model errors develop over long interactions. We find that externally increased task pressure can sharply raise the probability of collapse, while naturally occurring pressure is only a modest predictor of later failure. Errors are temporally correlated rather than independent, and their propagation changes with dependency depth while remaining bounded in finite tasks. We also find that locally valid actions can continue after the agent's overall task state has become incorrect, that error structure depends on task topology and horizon, and that belief states in embodied tasks are shallow rather than strongly trapped in incorrect states. Larger horizons allow larger error cascades, but their scale remains constrained by the task and the model's capability. Overall, we observe power-law behavior, bounded divergence consistent with weak chaos, and metastable belief dynamics across long-horizon agent interactions. We call this perspective \emph{world model science}: the empirical study of self organized criticality in long-horizon agent world models.
\end{abstract}


\newcounter{wmFigureDescriptionIndex}
\newcommand{\wmApplyFigureDescription}{%
  \stepcounter{wmFigureDescriptionIndex}%
  \ifcase\value{wmFigureDescriptionIndex}%
  \or\Description{Recording, alignment, and diagnostic workflow. Agent observations, actions, and environment responses are converted into agent-implied and benchmark-grounded states, aligned over time, and compared. The resulting errors are evaluated through error clustering, temporal dependence, graph-conditioned propagation, and belief dynamics. A lower strip illustrates slowly accumulated stress, cascade size, and recovery.}%
  \or\Description{Two schematic capability boundaries. The upper panel plots a rising response curve against input size and marks a comprehension boundary. The lower panel shows valid Game-of-Life outputs across grid sizes, followed by an invalid-output region beyond the supported output size.}%
  \or\Description{Two series share an injected-stress axis: collapse rate and mean avalanche size. Collapse rate rises sharply with stress, and mean avalanche size also increases. A highlighted annotation reports an AUROC of 0.979.}%
  \or\Description{Two panels. The left horizontal bars compare the local-global gap across GAIA step types; the largest gap is at intermediate conclusions. The right panel compares reward and collapse-rate bars with a recovery-time line across five verification policies in Retail.}%
  \or\Description{Three panels compare Kolmogorov-Smirnov distances from the identity condition under task-preserving surface changes in StatefulPuzzle-SOC, HotpotQA-RAG, and Retail. Bars report the observed distances, and horizontal dashed lines show the null thresholds.}%
  \or\Description{Two panels. The left shows error-autocorrelation exponents with uncertainty bars across four interaction horizons and a horizontal reference at one. The right compares exact-match and F1 bars with an error-autocorrelation line across memory and retrieval settings.}%
  \or\Description{Two panels. The left plots maximum and mean avalanche size against interaction horizon on logarithmic axes. The right plots fractal dimension with uncertainty bars against grid size and shades the region where valid trajectories are unavailable.}%
  \or\Description{Two panels compare error propagation across dependency depths. The left plots the AIC difference between exponential and power-law fits, favoring the exponential fit at depth one and the power-law fit at larger depths. The right shows the fraction of trajectory pairs favoring the power-law fit and a 50 percent reference line.}%
  \or\Description{Three panels show fractal dimensions for HotpotQA retrieval topologies, ALFWorld task layouts, and Game-of-Life grid sizes. The first two panels use bars; the third uses a line with uncertainty bars.}%
  \or\Description{Two panels assess clustering of Retail prompt regimes. The left plots silhouette scores against the number of clusters, with the highest score at two clusters. The right compares within-cluster spread, between-cluster distance, and their separation ratio.}%
  \or\Description{Two side-by-side panels assess Retail stress prediction. The left compares AUROC for the reward-error and concurrent fidelity-drop labels. The right compares the full stress score and its individual components against the independent reward-error label. Dashed lines indicate chance performance.}%
  \or\Description{Two panels compare mean reward across intervention policies. ALFWorld reaches its highest mean reward under exploration-heavy or memory-heavy settings, while Retail reaches its highest mean reward under no intervention or high verification. Stars mark the best-performing settings.}%
  \fi
}
\AtBeginEnvironment{figure}{\wmApplyFigureDescription}
\AtBeginEnvironment{figure*}{\wmApplyFigureDescription}

\begin{document}
\maketitle

\section{Introduction}

Long-horizon LLM agents must maintain and update a world model over extended sequences of observations, actions, decisions, and tool calls. Tool-using agents must track database states, policy constraints, previous calls, and their consequences when interacting with external tools. Examples include $\tau$-bench~\citep{yao2024taubench}, API-Bank~\cite{li2023apibank}, ToolLLM~\citep{qin2024toolllm}, and StableToolBench~\citep{guo2024stabletoolbench}. Retrieval agents must integrate and update evidence across multiple reasoning steps. HotpotQA provides a representative setting in which answering a question requires reasoning over information from multiple documents \citep{yang2018hotpotqa}. Embodied agents must maintain room states, inventory states, and task progress while interacting with an environment. ALFWorld evaluates agents on such multi-step embodied tasks \citep{shridhar2020alfworld}. As interactions become longer, agents must remember more information and rely on more previous decisions, making errors more likely to affect later steps \citep{liu2023agentbench,wang2026horizon}. As a result, an agent may continue to take valid actions even after it has lost an accurate understanding of the overall task \citep{chung2026where,cuadron2025saber}.

\begin{figure*}[!ht]
\centering
\includegraphics[width=.75\linewidth]{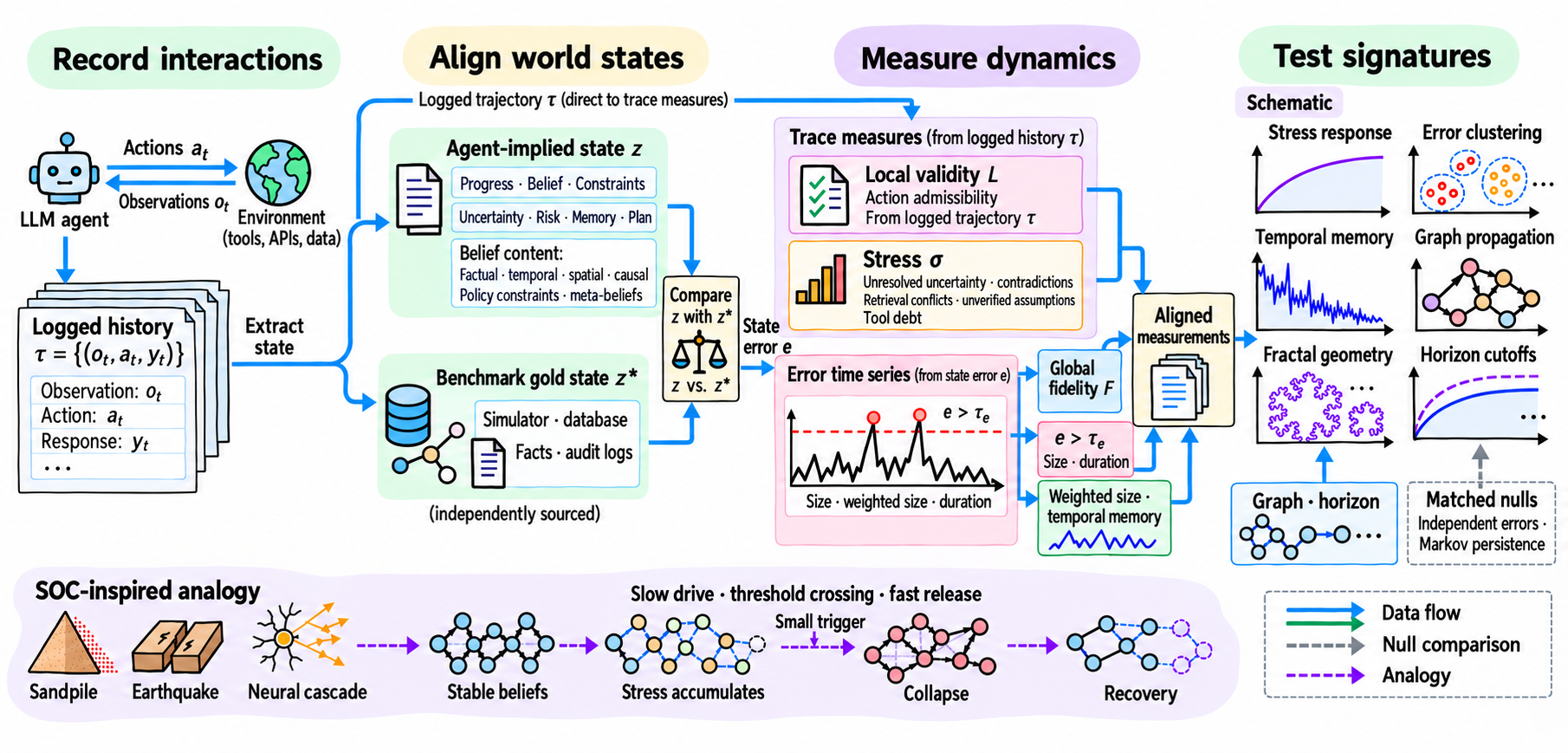}

\caption{\textbf{Overview of our framework.} We align agent-implied states with benchmark-grounded states and use the resulting trajectories to measure and test the dynamics of long-horizon agent world models.}
\label{fig:intuition}

\end{figure*}

Existing evaluations largely measure whether an agent succeeds or whether individual actions are valid, but reveal much less about the mechanisms that lead to failure. Recent work has begun to analyze failures along complete agent trajectories. HORIZON, for example, studies where agents fail as task horizons increase and attributes failures to different categories across web, operating-system, database, and embodied tasks \citep{wang2026horizon}. Other work introduces checkpoints into long-horizon security tasks to determine whether an agent fails before or after reaching the state needed for a later decision \citep{shao2026beyond}. Related trajectory-level evaluations further show that identifying the exact step or component responsible for a long-horizon failure remains difficult \citep{wang2026hintbench,chen2026trajectory}. However, these studies primarily locate or categorize failures rather than explain the dynamics that produce them. We still lack a clear understanding of how errors develop within an agent's world model, how an early error changes later behavior, and why some agents recover while others suddenly fail. We therefore study the mechanisms of long-horizon agent failure from three dynamical perspectives: \textbf{self-organized criticality, weak chaos, and metastable belief dynamics}.

Self-organized criticality provides one perspective on abrupt failure. In many complex systems, small changes can accumulate before producing events at much larger scales, from sandpile and crackling-noise systems to cellular automata \citep{bak1987self,bak1997nature,sethna2001crackling,stanley1996anomalous,langton1990computation}. Similar behavior has been studied in neural systems, where neural avalanches exhibit cascades and scale-dependent dynamics near critical regimes \citep{beggs2003neuronal,chialvo2010emergent,shew2009neuronal,shew2013functional,plenz2021self}. Long-horizon agents may exhibit similar dynamics: small world-model errors can accumulate across interactions and eventually lead to much larger failures. From this perspective, we experimentally study whether agent errors form correlated cascades, exhibit power-law behavior, and change systematically as the interaction horizon grows.

Weak chaos provides a second perspective on how small errors propagate over time. In dynamical systems, small differences can grow as the system evolves, and the resulting divergence characterizes different dynamical regimes \citep{strogatz2018nonlinear,ott2002chaos}. Long-horizon agents may exhibit similar behavior because an early world-model error can affect decisions many steps later. This raises the question of how dependency depth changes error propagation and whether divergence continues to grow or eventually becomes bounded. Metastable belief dynamics provides a third perspective on how agents move between different beliefs about the world. Metastable systems can remain in relatively stable states before transitioning to others \citep{friston1997transients,tognoli2014metastable,brinkman2022metastable,rossi2025dynamical}. Long-horizon agents may similarly remain in correct or incorrect belief states before switching or recovering, raising questions about how long these states persist and how transitions occur. To examine these dynamics, we reconstruct the world state implied by an agent's interaction history and compare it with the benchmark-grounded state over time.

We record agent interactions, reconstruct the world state implied by each agent, and compare it with the benchmark-grounded state to track world-model errors over time. We conduct extensive experiments on controlled puzzles, tool-use tasks, embodied environments, multi-hop retrieval, general-assistant reasoning, and Game of Life, including StatefulPuzzle-SOC, $\tau$-bench Retail and Airline \citep{yao2024taubench}, GAIA Level-1 \citep{mialon2023gaia}, ALFWorld \citep{shridhar2020alfworld}, HotpotQA-RAG \citep{yang2018hotpotqa}, and Game of Life \citep{gardner1970life}. Figure~\ref{fig:intuition} summarizes our framework. We find that world-model errors are temporally correlated and form power-law error cascades, with larger horizons supporting larger cascades. Error propagation changes with dependency depth but remains bounded, consistent with weak chaos. Agents can also continue taking locally valid actions after their overall task state becomes incorrect, while belief states in embodied tasks are shallow and recoverable rather than strongly trapped.

These findings lead us to propose \emph{world model science}: a framework for understanding the dynamics of long-horizon agent world models through three complementary views. Self-organized criticality describes how errors accumulate and form cascades, weak chaos describes how small errors propagate while remaining bounded, and metastable belief dynamics describes how agents persist in and transition between different belief states. Together, these views provide a unified way to study how agent world models change during long-horizon interactions.

Our contributions are summarized as follows:
\begin{enumerate}
\item We introduce \emph{world model science} to study the internal mechanisms and natural evolution of long-horizon LLM agents through self-organized criticality, weak chaos, and metastable belief dynamics.
\item We develop a unified framework for reconstructing agent-implied world states and testing these three dynamics against benchmark-grounded states across long-horizon interactions.
\item We conduct extensive tests across controlled and benchmark environments, revealing power-law error cascades, bounded error propagation consistent with weak chaos, and persistent but recoverable belief states.
\end{enumerate}

\section{Related Work}
\label{sec:related_work}

\textbf{Agent evaluation, state tracking, and long-horizon reliability.}
Interactive language agents combine reasoning, acting, memory, and feedback over multi-step interactions \citep{wei2022chain,yao2023react,shinn2023reflexion,madaan2023selfrefine,schick2023toolformer}. Benchmarks evaluate tool use and policy compliance \citep{li2023apibank,qin2024toolllm,guo2024stabletoolbench,yao2024taubench}, extended web and software tasks \citep{zhou2024webarena,jimenez2024swebench}, and failures within agent trajectories \citep{cemri2025why,zhu2025where,zhu2026dissecting}. Related work further examines semantic state tracking, mutating actions, toolchain propagation, hallucination cascades, and epistemic calibration \citep{chung2026where,cuadron2025saber,xiong2025butterfly,jamshidi2026hallucination,wang2026planning}, as well as accumulated errors, behavioral drift, and performance degradation over long horizons \citep{kwa2025measuring,sinha2025illusion,rath2026agent,gurram2026evaluating}. We focus on how benchmark-grounded world-state errors evolve across trajectories and task dependencies.

\textbf{Criticality-inspired diagnostics.}
Self-organized criticality and crackling-noise theory characterize event-size distributions, finite-size scaling, temporal dependence, and invariance \citep{bak1987self,bak1997nature,sethna2001crackling,stanley1971introduction}, with related ideas applied to neural cascades, dynamic range, and long-range temporal structure \citep{beggs2003neuronal,shew2009neuronal,chialvo2010emergent,shew2013functional,plenz2021self}. We bring these statistical tools to long-horizon agents to study error cascades, temporal memory, propagation, and finite-size effects in world-model trajectories.

\begin{table*}[!ht]
\centering
\small
\setlength{\tabcolsep}{6pt}
\renewcommand{\arraystretch}{1.12}
\caption{\textbf{Comparison with closely related work.}
$\checkmark$ indicates that the corresponding phenomenon is explicitly studied or measured.}
\label{tab:related_work_novelty}

\begin{tabular}{lcccccc}
\toprule
\textbf{Related work}
& \textbf{\shortstack{State\\Error}}
& \textbf{\shortstack{Temporal\\Dependence}}
& \textbf{\shortstack{Error\\Cascades}}
& \textbf{\shortstack{Structural\\Propagation}}
& \textbf{\shortstack{Horizon\\Scaling}}
& \textbf{\shortstack{Belief\\Recovery}} \\
\midrule

World-state evaluation
\citep{vafa2024evaluating,harang2025tracking}
& \checkmark & \xmark & \xmark & \xmark & \xmark & \xmark \\

Trajectory diagnosis
\citep{chung2026where,wang2026horizon,cuadron2025saber}
& \checkmark & \checkmark & \xmark & \xmark & \checkmark & \xmark \\

Error propagation
\citep{sinha2025illusion,xiong2025butterfly,gurram2026evaluating}
& \checkmark & \checkmark & \checkmark & \checkmark & \xmark & \xmark \\

Hallucination cascades
\citep{jamshidi2026hallucination}
& \xmark & \checkmark & \checkmark & \xmark & \xmark & \xmark \\

Duration and persistence
\citep{kwa2025measuring,laban2025lost}
& \xmark & \checkmark & \xmark & \xmark & \checkmark & \xmark \\

Critical dynamics
\citep{vock2025critical}
& \xmark & \checkmark & \checkmark & \checkmark & \xmark & \xmark \\

World-model collapse
\citep{song2026worldmodelcollapse}
& \checkmark & \checkmark & \xmark & \checkmark & \checkmark & \xmark \\

\midrule

\rowcolor{gray!15}
\textbf{World Model Science (ours)}
& \checkmark
& \checkmark
& \checkmark
& \checkmark
& \checkmark
& \checkmark \\

\bottomrule
\end{tabular}

\end{table*}

\textbf{Positioning relative to prior work.}
Prior work has separately studied world-state fidelity, trajectory failures, error propagation, interaction duration, and critical dynamics \citep{vafa2024evaluating,harang2025tracking,chung2026where,wang2026horizon,sinha2025illusion,xiong2025butterfly,kwa2025measuring,vock2025critical}. We study world-state errors as an evolving process, jointly examining their temporal dependence, cascade behavior, structural propagation, horizon scaling, and recovery. Unlike the closest precursor \citep{song2026worldmodelcollapse}, which focuses on where world-model collapse begins, we characterize how the resulting error dynamics evolve after onset. Table~\ref{tab:related_work_novelty} summarizes these distinctions.

\section{Preliminaries: Measuring Agent World States}\label{sec:preliminaries}

We model each agent run as a finite-horizon partially observable interaction with observations, actions, and environment responses \citep{lovejoy1991pomdp,sutton2018reinforcement}. Our analysis does not require access to model activations. It uses only the interaction log and benchmark-specific measurements fixed before evaluation.

\begin{definition}[Agent trajectory]
For task $i$ with horizon $H_i$, let $o_{i,t}$, $a_{i,t}$, and $y_{i,t}$ denote the observation, agent action, and environment response at step $t$, respectively. The complete interaction trajectory is
\begin{equation}
\tau_i=\{(o_{i,t},a_{i,t},y_{i,t})\}_{t=1}^{H_i}.
\label{eq:trajectory}
\end{equation}
The interaction history through step $t$ is
\begin{equation}
h_{i,t}=(o_{i,1:t},a_{i,1:t},y_{i,1:t}).
\label{eq:history}
\end{equation}
\end{definition}

\begin{definition}[Agent-implied and gold world states]
For each benchmark substrate $s$, let $\mathcal Z_s$ denote its world-state space. A fixed extractor $\phi_s$ maps the interaction history to the world state implied by the agent:
\begin{equation}
z_{i,t}=\phi_s(h_{i,t})\in\mathcal Z_s.
\label{eq:agent_state}
\end{equation}
A benchmark-specific extractor $\phi_s^\star$ independently constructs the gold world state:
\begin{equation}
z_{i,t}^\star=\phi_s^\star(h_{i,t})\in\mathcal Z_s.
\label{eq:gold_state}
\end{equation}
\end{definition}
The agent-implied state $z_{i,t}$ is inferred from the interaction history, while $z_{i,t}^\star$ is obtained from benchmark information such as simulator states, database states, supporting facts, or audit logs. Their benchmark-specific definitions are given in Appendix~\ref{app:operationalization}. Thus, $z_{i,t}$ represents the world state implied by the agent, whereas $z_{i,t}^\star$ represents the state supported by the benchmark.

\begin{definition}[World-model error and fidelity]
For component $k$, let $\omega_k\geq0$ be a fixed weight with $\sum_k\omega_k=1$. The world-model error is
\begin{equation}
e_{i,t}=\left(\sum_k\omega_k\left\lVert z_{i,t}^{(k)}-z_{i,t}^{\star(k)}\right\rVert^2\right)^{1/2},
\label{eq:world_model_error}
\end{equation}
and world-state fidelity is
\begin{equation}
F_{i,t}=1-e_{i,t}.
\label{eq:fidelity}
\end{equation}
\end{definition}

The error sequence $e_{i,1},\ldots,e_{i,H_i}$ describes how the agent's world-model error evolves over the trajectory.

\begin{definition}[Local--global gap]
For each substrate $s$, let $\mathcal V_s(h_{i,t-1},y_{i,t})$ denote the set of actions that are valid at step $t$ given the interaction history before the action and the resulting environment response. We define local action validity as
\begin{equation}
L_{i,t}=\mathbf{1}\!\left[a_{i,t}\in\mathcal V_s(h_{i,t-1},y_{i,t})\right].
\label{eq:local_validity}
\end{equation}
The local--global gap is
\begin{equation}
\Delta_{i,t}^{LG}=L_{i,t}-F_{i,t}.
\label{eq:local_global_gap}
\end{equation}
\end{definition}
Thus, $L_{i,t}=1$ when the executed action $a_{i,t}$ is locally valid and $L_{i,t}=0$ otherwise. A positive gap indicates that the current action is locally valid even though the agent's world state is inaccurate.

\begin{definition}[Error avalanche]
Let $\tau_e\in(0,1)$ be a fixed error threshold. For the $j$-th error avalanche in trajectory $i$, let $s_{i,j}$ and $r_{i,j}$ denote its starting and ending steps. The avalanche is the maximal consecutive interval
\begin{equation}
\mathcal A_{i,j}=\{t:s_{i,j}\leq t\leq r_{i,j},\ e_{i,t}>\tau_e\},
\label{eq:error_avalanche}
\end{equation}
where $e_{i,s_{i,j}-1}\leq\tau_e$ if $s_{i,j}>1$ and $e_{i,r_{i,j}+1}\leq\tau_e$ if $r_{i,j}<H_i$. Its size and duration are
\begin{equation}
A_{i,j}=\sum_{t=s_{i,j}}^{r_{i,j}}e_{i,t}, \qquad D_{i,j}=r_{i,j}-s_{i,j}+1.
\label{eq:avalanche_size_duration}
\end{equation}
\end{definition}

\begin{assumption}[Frozen measurement protocol]\label{ass:frozen}
For each substrate $s$, the state extractors $\phi_s$ and $\phi_s^\star$, component weights $\omega_k$, and error threshold $\tau_e$ are fixed before outcome analysis, with
\begin{equation}
\omega_k\geq0,\qquad \sum_k\omega_k=1,\qquad \tau_e\in(0,1).
\label{eq:frozen_measurement}
\end{equation}
\end{assumption}
Appendix~\ref{app:operationalization} specifies their substrate-specific operationalization.

We restrict dynamical analyses to capability-matched trajectories that satisfy the independent validity requirements of each substrate. If a model produces no valid trajectories in a given regime, we treat that regime as outside the model's supported capability range rather than interpreting the absence of a dynamical pattern as evidence against that pattern.

\section{Experimental Protocol}

We conduct a fixed suite of experiments under a common model interface with temperature zero and a fixed seed. The experiments cover StatefulPuzzle~\cite{toshniwal2022chess} for controlled interventions on error accumulation, stress, horizon, dependency depth, temporal dependence, and finite-size scaling; $\tau$-bench Retail \citep{yao2024taubench} for policy-constrained API use, naturally occurring error precursors, and intervention analysis; $\tau$-bench Airline \citep{yao2024taubench} for policy-constrained API use and local--global validity gaps; GAIA Level-1 \citep{mialon2023gaia} for assistant-style evidence gathering and local--global state divergence; ALFWorld \citep{shridhar2020alfworld} for embodied state tracking, belief-state transitions, and graph-conditioned error geometry; HotpotQA-RAG \citep{yang2018hotpotqa} for multi-hop retrieval, temporal error structure, and graph-conditioned geometry; and Game of Life \citep{gardner1970life} for controlled local-rule dynamics, horizon scaling, and capability-boundary analysis.

Our protocol measures agent dynamics throughout each trajectory rather than relying only on terminal success. This distinction is motivated by prior work showing that trajectory-level state drift, mutating actions, and reasoning errors can reveal failures that final outcomes alone do not capture \citep{chung2026where,cuadron2025saber,cemri2025why,zhu2025where,zhu2026dissecting}. It is also important to distinguish executable actions from globally correct behavior: tool and interactive-agent benchmarks show that syntactic validity, policy compliance, environment state, and task success need not coincide \citep{li2023apibank,qin2024toolllm,guo2024stabletoolbench,yao2024taubench,zhou2024webarena,jimenez2024swebench}. We therefore record the intermediate world states defined in the previous section and evaluate their evolution over the complete trajectory.

Figure~\ref{fig:intuition} summarizes the measurement pipeline. Each interaction trace is aligned with its benchmark-grounded world state to obtain the error sequence $e_{i,t}$, fidelity $F_{i,t}$, local--global gap $\Delta_{i,t}^{LG}$, and error avalanches. This separation allows us to distinguish local action validity from global world-state fidelity and to detect cases in which an agent continues to take valid actions after its world model has become inaccurate.

Table~\ref{tab:cross_substrate} summarizes the evidence for each diagnostic across substrates, while Table~\ref{tab:evidence_map} links these diagnostics to their corresponding analyses. Sections~\ref{sec:collapse_results} and~\ref{sec:structure_results} present the main results on error accumulation, local--global gaps, temporal dependence, error propagation, geometry, and horizon scaling. Appendix~\ref{app:moved_results} provides additional boundary and directional analyses.

\begin{table*}[!ht]
\centering
\small
\renewcommand{\arraystretch}{1.35}
\setlength{\tabcolsep}{6pt}

\definecolor{support}{RGB}{218,242,224}
\definecolor{partial}{RGB}{232,222,250}
\definecolor{boundary}{RGB}{255,232,199}
\definecolor{alternative}{RGB}{255,242,204}
\definecolor{capacity}{RGB}{218,231,250}
\definecolor{nottest}{RGB}{242,242,242}

\resizebox{0.8\textwidth}{!}{%
\begin{tabular}{@{}lccccccc@{}}
\toprule
\textbf{Diagnostic} &
\shortstack{\textbf{Stateful}\\\textbf{Puzzle}} &
\shortstack{\textbf{$\tau$-bench}\\\textbf{Retail}} &
\shortstack{\textbf{$\tau$-bench}\\\textbf{Airline}} &
\shortstack{\textbf{GAIA}\\\textbf{L1}} &
\textbf{ALFWorld} &
\shortstack{\textbf{HotpotQA}\\\textbf{RAG}} &
\shortstack{\textbf{Game}\\\textbf{of Life}} \\
\midrule

Small-error trigger and collapse
& \cellcolor{support}\textbf{S}
& \cellcolor{support}\textbf{S}
& \cellcolor{nottest}$\times$
& \cellcolor{nottest}$\times$
& \cellcolor{nottest}$\times$
& \cellcolor{nottest}$\times$
& \cellcolor{nottest}$\times$ \\

Belief-state persistence and recovery
& \cellcolor{nottest}$\times$
& \cellcolor{alternative}\textbf{A}
& \cellcolor{nottest}$\times$
& \cellcolor{nottest}$\times$
& \cellcolor{alternative}\textbf{A}
& \cellcolor{nottest}$\times$
& \cellcolor{nottest}$\times$ \\

Local action vs.\ global state
& \cellcolor{nottest}$\times$
& \cellcolor{nottest}$\times$
& \cellcolor{support}\textbf{S}
& \cellcolor{support}\textbf{S}
& \cellcolor{nottest}$\times$
& \cellcolor{nottest}$\times$
& \cellcolor{nottest}$\times$ \\

Stability under surface perturbations
& \cellcolor{support}\textbf{S}
& \cellcolor{nottest}$\times$
& \cellcolor{nottest}$\times$
& \cellcolor{nottest}$\times$
& \cellcolor{nottest}$\times$
& \cellcolor{support}\textbf{S}
& \cellcolor{nottest}$\times$ \\

Error accumulation over time
& \cellcolor{nottest}$\times$
& \cellcolor{support}\textbf{S}
& \cellcolor{nottest}$\times$
& \cellcolor{nottest}$\times$
& \cellcolor{nottest}$\times$
& \cellcolor{nottest}$\times$
& \cellcolor{nottest}$\times$ \\

Long-range temporal dependence
& \cellcolor{partial}\textbf{P}
& \cellcolor{nottest}$\times$
& \cellcolor{nottest}$\times$
& \cellcolor{nottest}$\times$
& \cellcolor{nottest}$\times$
& \cellcolor{support}\textbf{S}
& \cellcolor{nottest}$\times$ \\

Error spread with dependency depth
& \cellcolor{support}\textbf{S}
& \cellcolor{nottest}$\times$
& \cellcolor{nottest}$\times$
& \cellcolor{nottest}$\times$
& \cellcolor{nottest}$\times$
& \cellcolor{nottest}$\times$
& \cellcolor{partial}\textbf{P} \\

Graph-conditioned error geometry
& \cellcolor{nottest}$\times$
& \cellcolor{nottest}$\times$
& \cellcolor{nottest}$\times$
& \cellcolor{nottest}$\times$
& \cellcolor{capacity}\textbf{C}
& \cellcolor{capacity}\textbf{C}
& \cellcolor{nottest}$\times$ \\

Stability across prompt variants
& \cellcolor{support}\textbf{S}
& \cellcolor{support}\textbf{S}
& \cellcolor{nottest}$\times$
& \cellcolor{nottest}$\times$
& \cellcolor{nottest}$\times$
& \cellcolor{support}\textbf{S}
& \cellcolor{nottest}$\times$ \\

Error scaling with interaction horizon
& \cellcolor{support}\textbf{S}
& \cellcolor{nottest}$\times$
& \cellcolor{nottest}$\times$
& \cellcolor{nottest}$\times$
& \cellcolor{nottest}$\times$
& \cellcolor{nottest}$\times$
& \cellcolor{capacity}\textbf{C} \\

Intervention-dependent behavior
& \cellcolor{nottest}$\times$
& \cellcolor{boundary}\textbf{B}
& \cellcolor{nottest}$\times$
& \cellcolor{nottest}$\times$
& \cellcolor{boundary}\textbf{B}
& \cellcolor{nottest}$\times$
& \cellcolor{nottest}$\times$ \\

\bottomrule
\end{tabular}%
}

\begin{center}
\footnotesize
\colorbox{support}{\strut\ \textbf{S}\ }\ Supporting
\quad
\colorbox{partial}{\strut\ \textbf{P}\ }\ Partial
\quad
\colorbox{boundary}{\strut\ \textbf{B}\ }\ Boundary
\quad
\colorbox{alternative}{\strut\ \textbf{A}\ }\ Alternative
\quad
\colorbox{capacity}{\strut\ \textbf{C}\ }\ Capacity-limited
\quad
\colorbox{nottest}{\strut\ $\times$\ }\ Not tested
\end{center}

\caption{\textbf{Cross-substrate diagnostic matrix.} Each cell reports the evidence for a diagnostic on a given substrate: supporting (S), partial (P), boundary (B), alternative (A), or capacity-limited (C).}
\label{tab:cross_substrate}

\end{table*}

\begin{table*}[!ht]
\centering
\scriptsize
\setlength{\tabcolsep}{3pt}
\resizebox{0.8\textwidth}{!}{%
\begin{tabular}{@{}p{.16\textwidth}p{.35\textwidth}p{.19\textwidth}p{.1\textwidth}@{}}
\toprule
Diagnostic block & Test & Evidence type & Location \\
\midrule
Stress collapse & External stress and avalanche nulls & Supporting evidence & Figure~\ref{fig:sigma} \\
Belief basins & Shallow-basin direction & Alternative explanation & Appendix~\ref{app:part2_moved} \\
Local--global gap & Local validity exceeds global fidelity & Supporting evidence & Figure~\ref{fig:lg} \\
Surface stability & Macro statistics under surface perturbations & Supporting evidence & Figure~\ref{fig:surface_stability} \\
Natural stress & Retail precursor prediction & Partial evidence & Figure~\ref{fig:stress_precursor_app} \\
Temporal memory & Spectral and valid-horizon DFA tests & Supporting evidence & Figure~\ref{fig:spectral} \\
Dependency transition & Bounded divergence with dependency depth & Supporting evidence & Figure~\ref{fig:chaos}; Table~\ref{tab:d_transition_app} \\
Error geometry & Graph-conditioned fractal dimension & Supporting evidence & Figure~\ref{fig:fractal_geometry} \\
Regime clustering & Prompt-variant macro clustering & Partial evidence & Figure~\ref{fig:universality_app} \\
Finite size & Avalanche cutoff scales with horizon & Supporting evidence & Figure~\ref{fig:fss} \\
Capability boundary & Valid-generation support in Game of Life & Capacity-limited result & Figures~\ref{fig:fss} and~\ref{fig:boundary} \\
Operating regime & Verification/exploration balance & Boundary evidence & Figure~\ref{fig:regimes_app} \\
\bottomrule
\end{tabular}}

\caption{\textbf{Diagnostic map.} Each block identifies the test used to characterize long-horizon world-model dynamics, the resulting evidence type, and the corresponding analysis.}
\label{tab:evidence_map}

\end{table*}

\begin{figure}[!ht]
\centering
\includegraphics[width=0.9\linewidth]{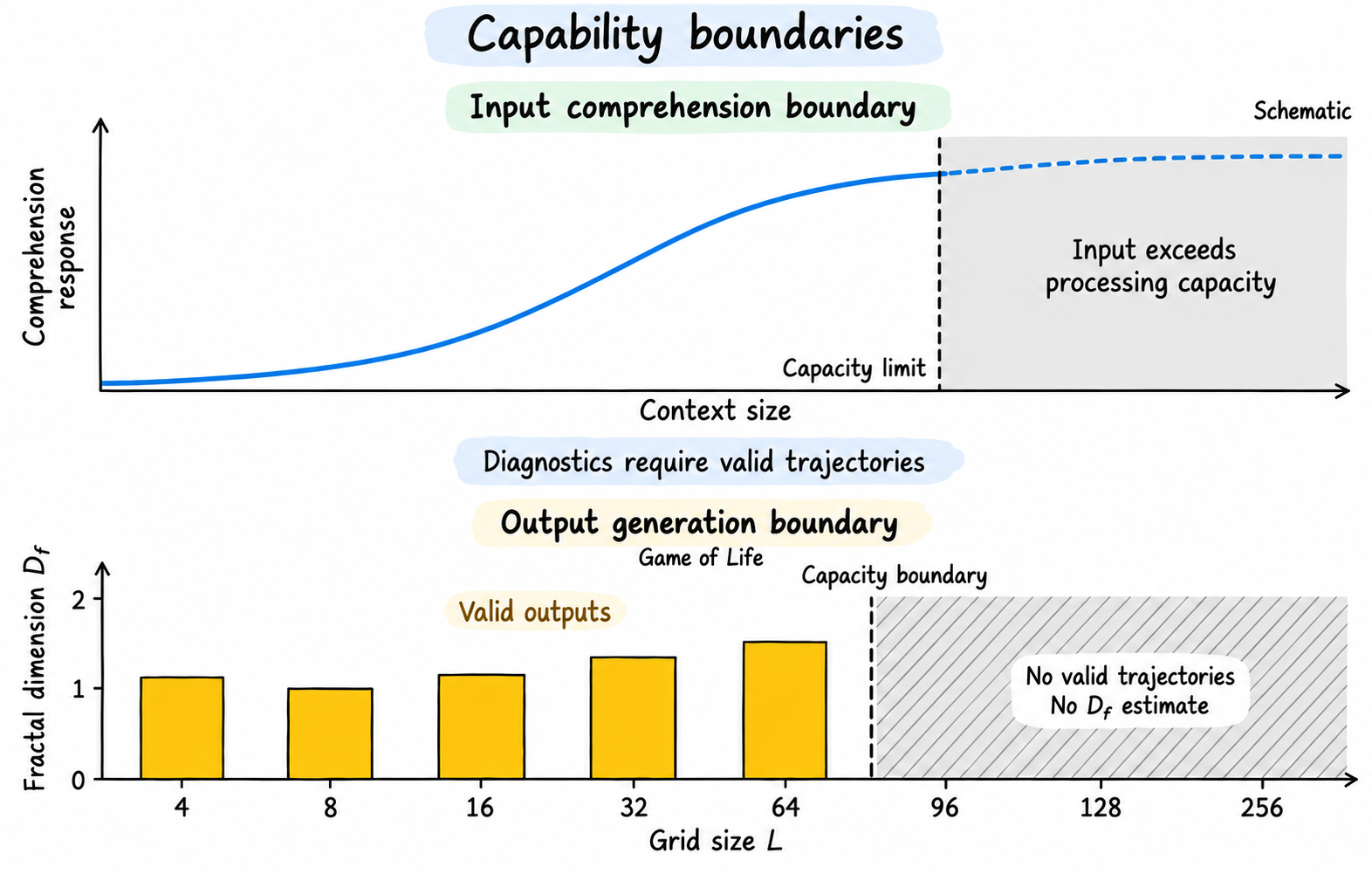}

\caption{\textbf{Capability boundaries.} Schematic input-comprehension boundary and Game-of-Life output-generation boundary.}
\label{fig:boundary}

\end{figure}

We use a common statistical protocol across all diagnostics. Confirmatory $p$-values are corrected using Benjamini--Hochberg false-discovery control at $q=0.05$ \citep{benjamini1995fdr}, and each diagnostic is compared with an appropriate null model. For independent per-step errors with probability $\varepsilon_0$, the horizon-$H$ baseline is $C_{\mathrm{ind}}(H)=1-(1-\varepsilon_0)^H$. Heavy-tail analyses follow finite-sample power-law procedures \citep{clauset2009power}, temporal-memory analyses use spectral and detrended-fluctuation methods \citep{peng1994dfa}, and graph-geometry analyses use box-counting estimands \citep{mandelbrot1982fractal,falconer2014fractal}.

\section{Results: Collapse Is Measurable and Not Independent Noise}\label{sec:collapse_results}
We first study the world-model error defined in Equation~\eqref{eq:world_model_error} and test whether its dynamics can be distinguished from independent per-step errors. We examine controlled stress, local--global gaps, task-preserving prompt changes, temporal dependence, and error propagation with dependency depth.

\subsection{Controlled Stress Produces Sharp Collapse}
\begin{figure}[!ht]
\centering
\includegraphics[width=0.9\linewidth]{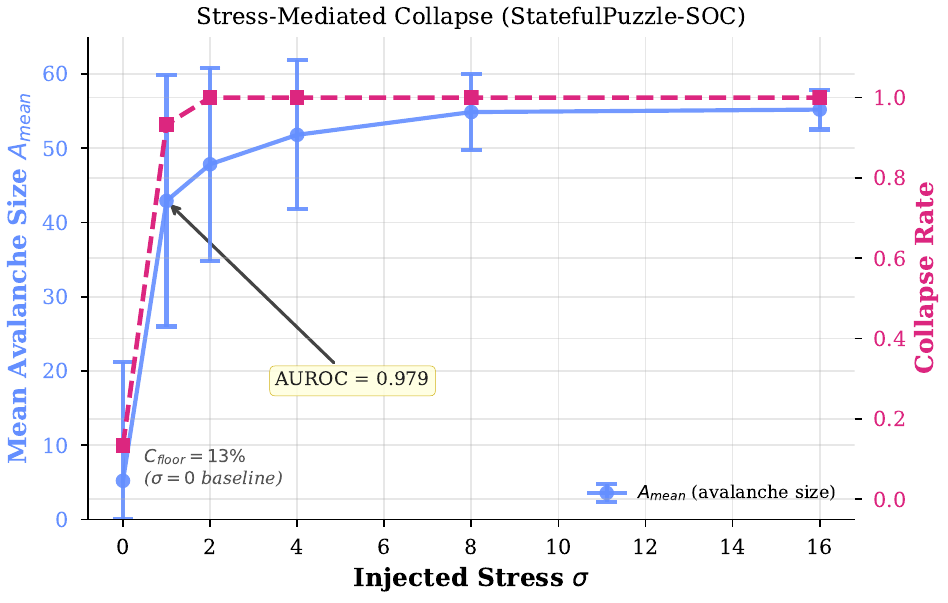}

\caption{\textbf{Collapse under injected stress in StatefulPuzzle-SOC.} Collapse rate and mean avalanche size across injected stress levels.}
\label{fig:sigma}

\end{figure}

We first test how externally injected stress affects the error avalanches defined in Equations~\ref{eq:error_avalanche}--\ref{eq:avalanche_size_duration}. In StatefulPuzzle-SOC at horizon $64$ and dependency depth one, collapse increases sharply with stress, and injected stress predicts collapse with AUROC $\mathbf{0.979}$. Figure~\ref{fig:sigma} shows collapse rate and mean avalanche size across stress levels.

\begin{figure*}[!ht]
\centering
\includegraphics[width=0.8\linewidth]{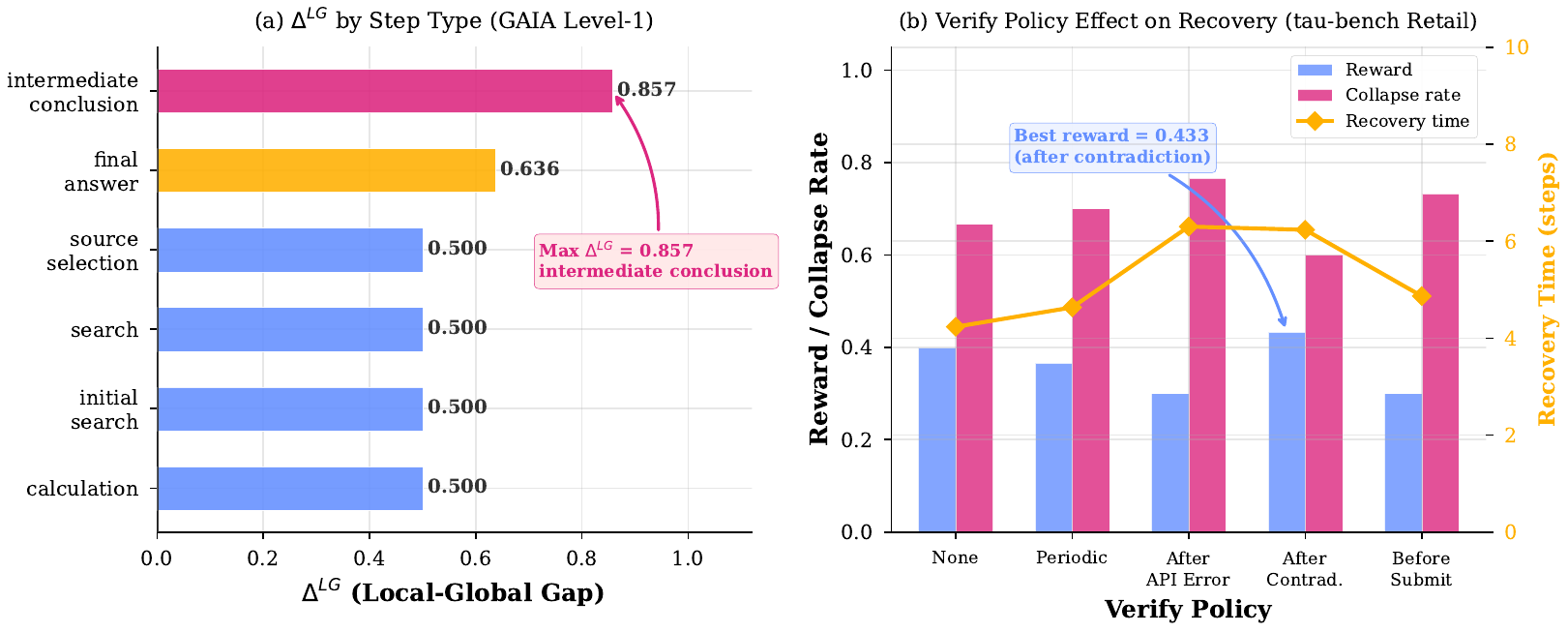}

\caption{\textbf{Local--global gap and verification.} Left: local--global gap by GAIA Level-1 step type. Right: reward, collapse rate, and recovery time under different verification policies in $\tau$-bench Retail.}
\label{fig:lg}

\end{figure*}

\begin{figure*}[!ht]
\centering
\includegraphics[width=0.8\linewidth]{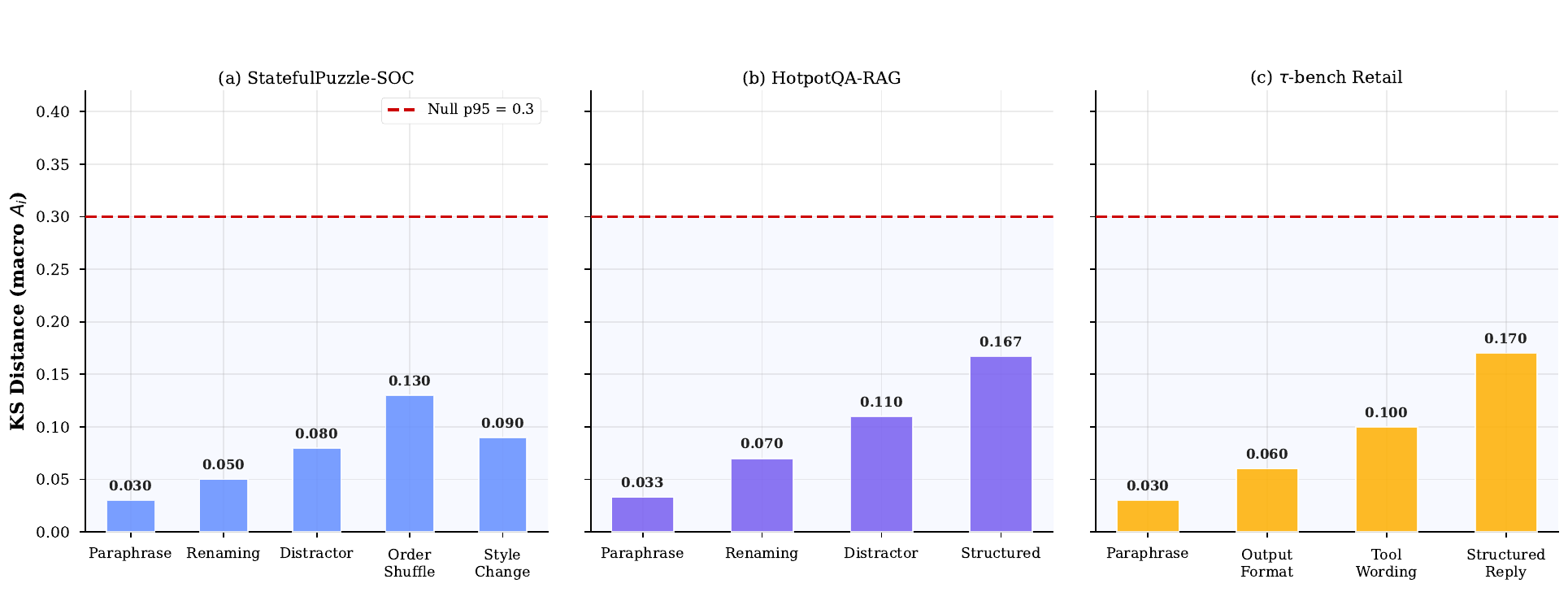}

\caption{\textbf{Stability under task-preserving changes.} KS distance from the identity baseline for prompt and formatting changes in StatefulPuzzle-SOC, HotpotQA-RAG, and $\tau$-bench Retail.}
\label{fig:surface_stability}

\end{figure*}

Retail provides a benchmark counterpart using the same avalanche definition in Equations~\ref{eq:error_avalanche}--\ref{eq:avalanche_size_duration}. Error avalanches are more temporally clustered than under an independent Bernoulli process, while truncated power-law, lognormal, and Markov models explain parts of the observed distribution.

\subsection{Local Actions Can Stay Valid After the World State Has Diverged}
We next measure the local--global gap defined in Equation~\eqref{eq:local_global_gap} to test whether locally valid behavior can continue after the agent's world state becomes incorrect. In GAIA Level-1, the largest gap occurs at intermediate conclusions, with $\Delta^{LG}=\mathbf{0.857}$. Figure~\ref{fig:lg} also compares verification policies in $\tau$-bench Retail.
\begin{figure*}[!ht]
\centering
\includegraphics[width=0.8\linewidth]{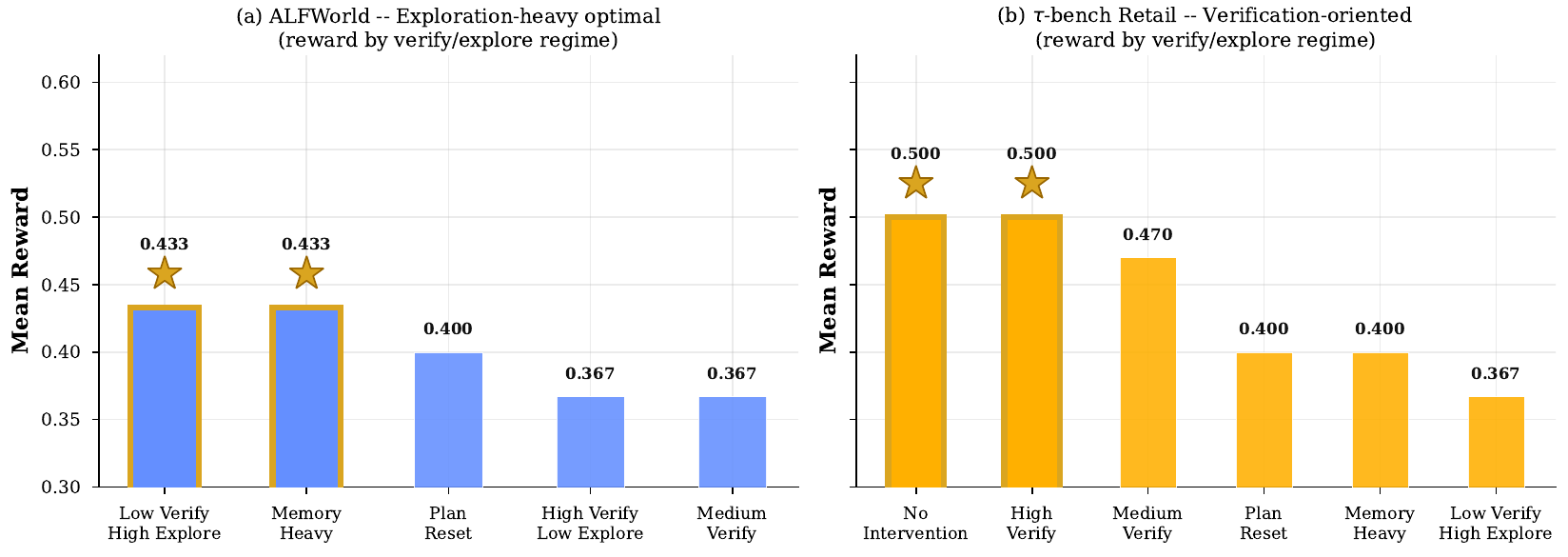}

\caption{\textbf{Intervention effects across tasks.} Mean reward under verification, exploration, memory, and planning settings in ALFWorld and $\tau$-bench Retail.}
\label{fig:regimes_app}

\end{figure*}

This gap shows why intermediate world-state tracking provides information beyond final success or local action validity.

\subsection{Collapse Statistics Are Stable Under Task-Preserving Changes}
We test whether task-preserving changes to wording, naming, ordering, and formatting alter the avalanche statistics defined in Equation~\eqref{eq:avalanche_size_duration}. Across StatefulPuzzle-SOC, HotpotQA-RAG, and $\tau$-bench Retail, all KS distances remain below the null threshold of $\mathbf{0.30}$. Figure~\ref{fig:surface_stability} reports the comparison with the identity baseline.

\subsection{Long-Horizon Errors Have Temporal Dependence}
We test for temporal dependence in the world-model error sequence $e_{i,1},\ldots,e_{i,H_i}$ defined by Equation~\eqref{eq:world_model_error}. In StatefulPuzzle, the error-autocorrelation exponent is $\alpha_e=\mathbf{1.52}$ at horizon $64$ and remains above the flicker-noise reference across the tested horizons. HotpotQA-RAG further shows that the temporal pattern changes with retrieval and memory conditions.

\begin{figure*}[!ht]
\centering
\includegraphics[width=0.8\linewidth]{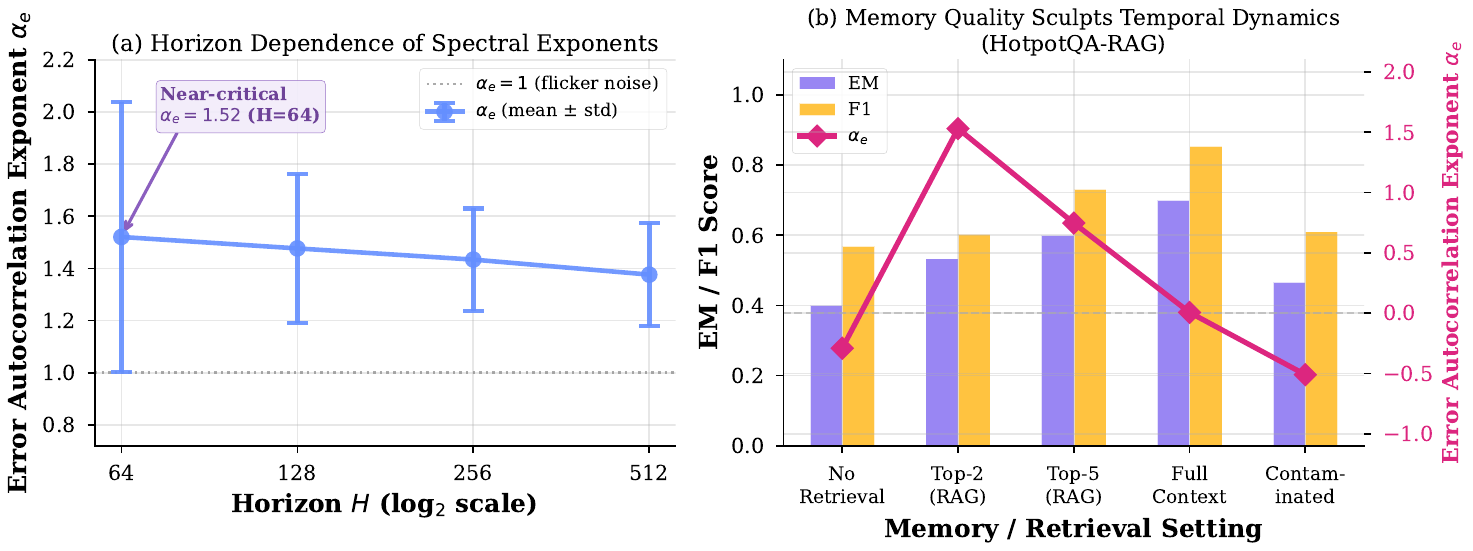}

\caption{\textbf{Temporal dependence in agent errors.} Left: error-autocorrelation exponent across StatefulPuzzle horizons. Right: EM, F1, and error-autocorrelation exponent across HotpotQA-RAG memory and retrieval settings.}
\label{fig:spectral}

\end{figure*}

\begin{figure*}[!ht]
\centering
\includegraphics[width=.75\linewidth]{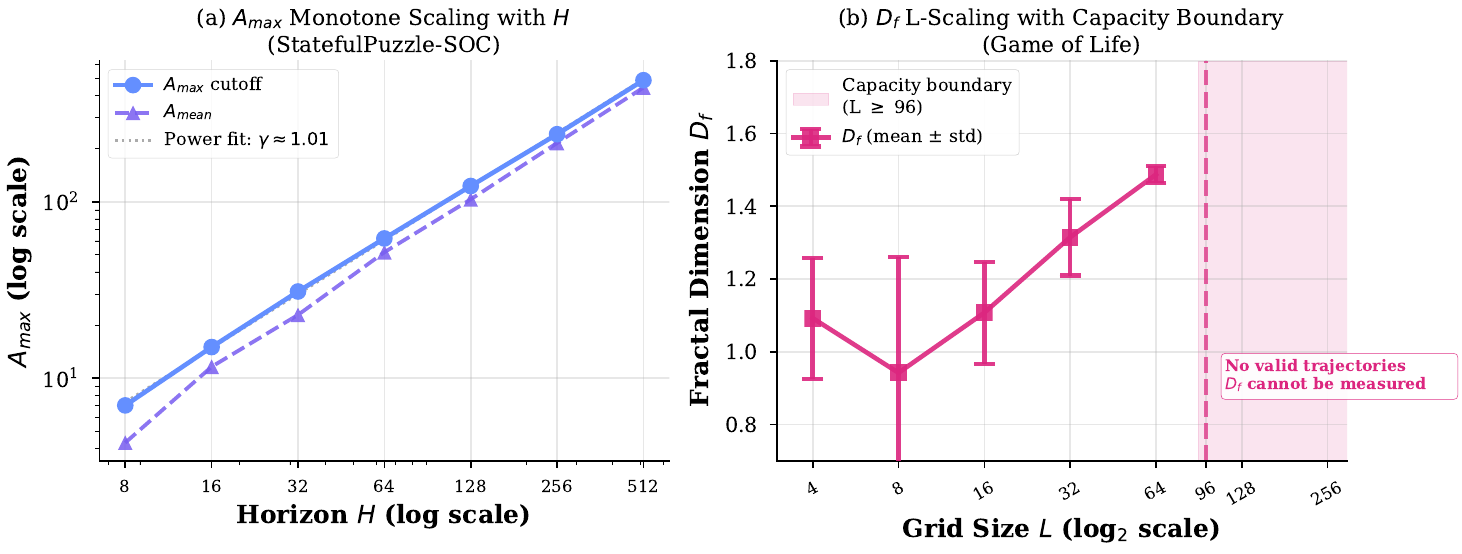}

\caption{\textbf{Scaling with horizon and grid size.} Left: maximum and mean avalanche size across StatefulPuzzle horizons. Right: fractal dimension across Game-of-Life grid sizes.}
\label{fig:fss}

\end{figure*}

\subsection{Dependency Depth Changes Error Propagation}
\begin{figure*}[!ht]
\centering
\includegraphics[width=0.8\linewidth]{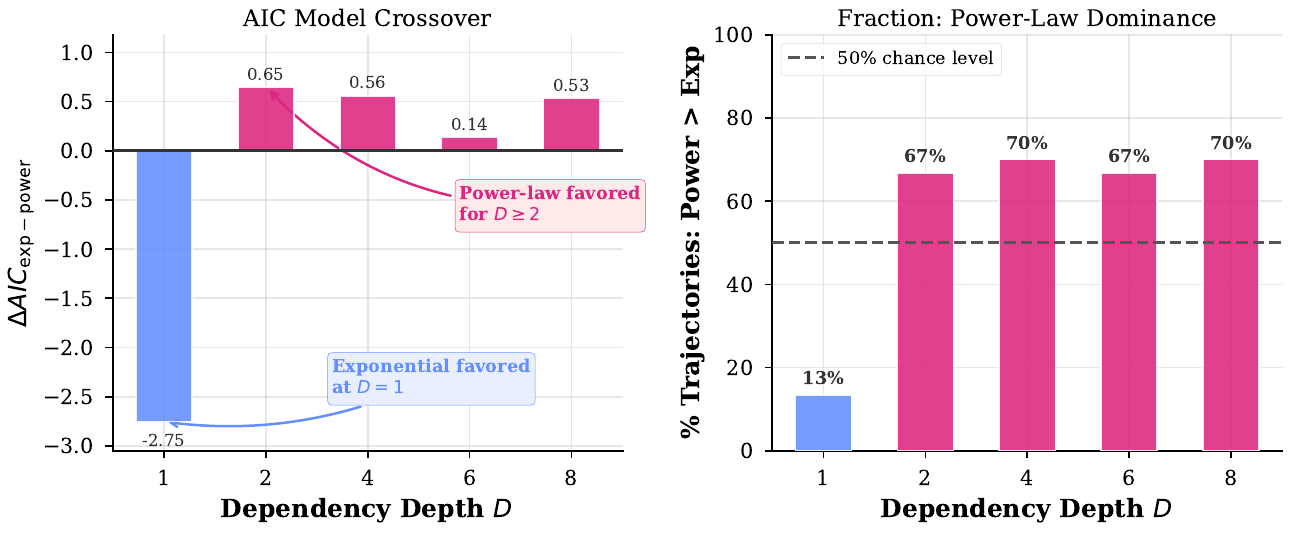}

\caption{\textbf{Error propagation across dependency depths.} Left: AIC difference between exponential and power-law fits. Right: fraction of trajectories favoring the power-law fit at each dependency depth.}
\label{fig:chaos}

\end{figure*}

We test how differences in the world-model error defined in Equation~\eqref{eq:world_model_error} propagate as dependency depth increases. At depth one, the exponential model is preferred, while for $D\geq2$ the power-law model is preferred by more than half of the paired trajectories; at $D=2$, this fraction is $\mathbf{67\%}$. Figure~\ref{fig:chaos} reports both the AIC comparison and trajectory-level model preference.

\begin{figure*}[!ht]
\centering
\includegraphics[width=.75\linewidth]{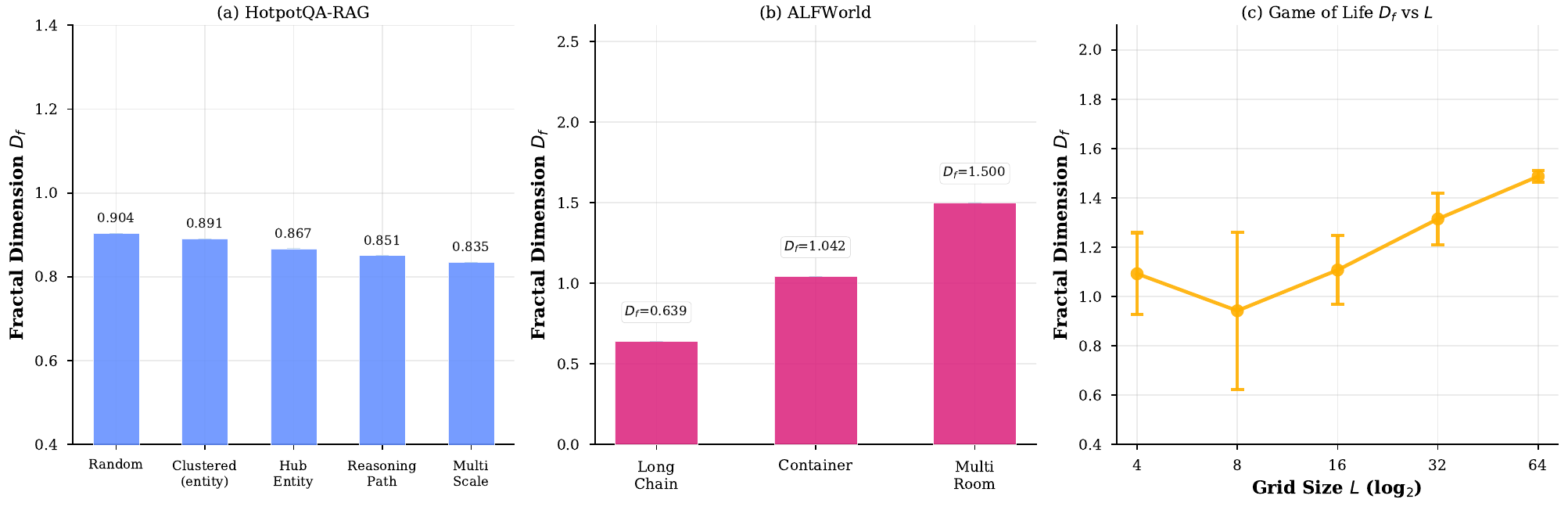}

\caption{\textbf{Error geometry across task structures.} Fractal dimension in HotpotQA-RAG, ALFWorld, and Game of Life.}
\label{fig:fractal_geometry}

\end{figure*}

\section{Results: Collapse Depends on Structure, Horizon, and Capability}\label{sec:structure_results}
We next study how the world-model error defined in Equation~\eqref{eq:world_model_error} changes with task structure, interaction horizon, and model capability.

\subsection{Error Geometry Depends on Task Structure}
We analyze the geometry of world-model errors defined in Equation~\eqref{eq:world_model_error} under different task structures. HotpotQA produces similar fractal dimensions across evidence structures, while ALFWorld varies more across task graphs, from $D_f=\mathbf{0.639}$ to $\mathbf{1.500}$. Game of Life provides a third comparison across grid sizes. Figure~\ref{fig:fractal_geometry} reports the three settings \citep{mandelbrot1982fractal,falconer2014fractal}.

\subsection{Larger Horizons Support Larger Error Cascades}
We vary the interaction horizon and measure the error avalanches defined in Equations~\ref{eq:error_avalanche}--\ref{eq:avalanche_size_duration} in StatefulPuzzle, together with error geometry across grid sizes in Game of Life. Maximum avalanche size increases with horizon, while Game of Life yields valid fractal-dimension estimates only through grid size $\mathbf{64}$. Figure~\ref{fig:fss} reports both scaling experiments.

Figure~\ref{fig:boundary} further separates two capability limits: understanding increasingly large inputs and generating valid outputs. Consistent with our capability-matched evaluation protocol, dynamical measurements are reported only when valid trajectories are available. The Game-of-Life experiment reaches an output-generation boundary beyond this range.

\begin{figure*}[!ht]
\centering
\includegraphics[width=0.8\linewidth]{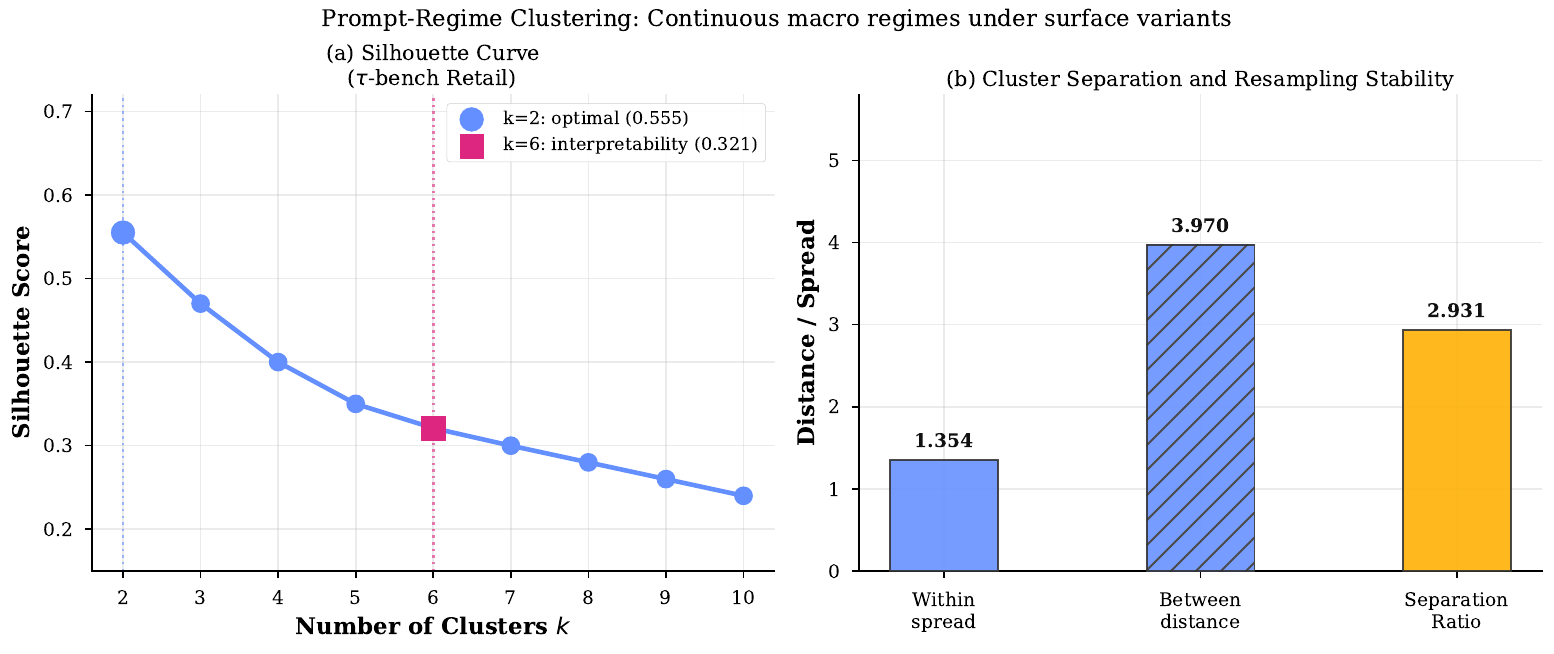}

\caption{\textbf{Prompt-regime clustering.} Left: silhouette score across numbers of clusters. Right: within-cluster spread, between-cluster distance, and separation ratio.}
\label{fig:universality_app}

\end{figure*}

\begin{figure*}[!ht]
\centering
\includegraphics[width=0.8\linewidth]{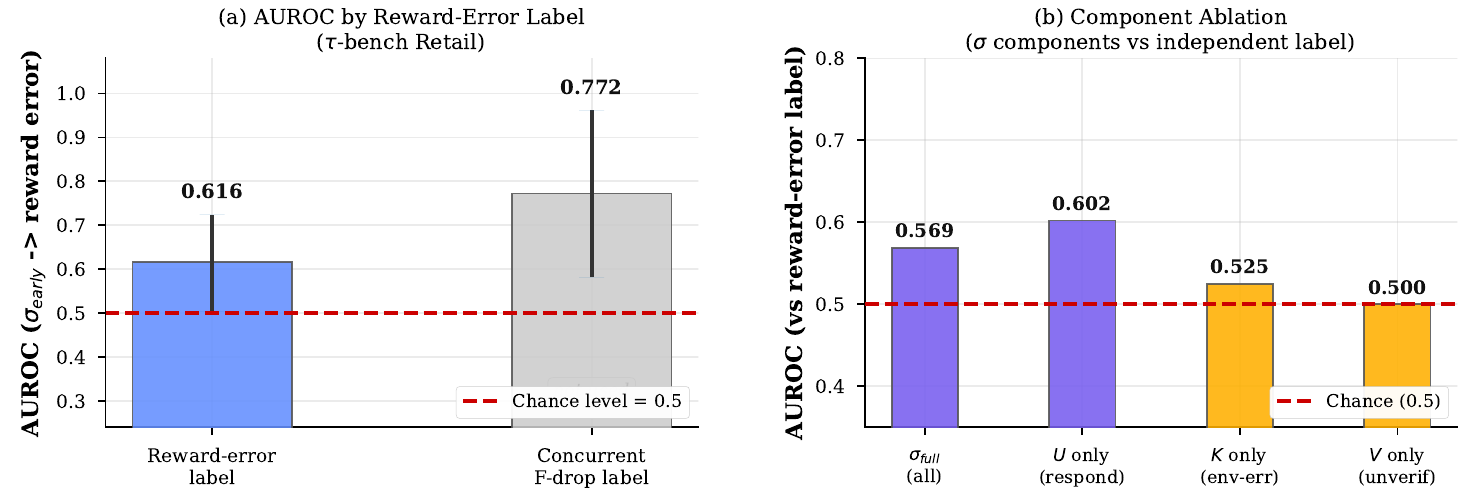}

\caption{\textbf{Retail stress prediction.} Top: AUROC under reward-error and concurrent fidelity-drop labels. Bottom: AUROC for individual stress components.}
\label{fig:stress_precursor_app}

\end{figure*}






\subsection{Boundary Tests for Stronger Claims}
We further test the robustness and limits of the dynamics measured from the world-model error in Equation~\eqref{eq:world_model_error} and world-state fidelity in Equation~\eqref{eq:fidelity}.

\paragraph{Natural Stress Is a Modest Predictor}
In $\tau$-bench Retail, we test whether early measurements derived from the agent's world-state trajectory predict a later reward error rather than a concurrent change in the fidelity defined in Equation~\eqref{eq:fidelity}. Early stress reaches an AUROC of $\mathbf{0.616}$. Figure~\ref{fig:stress_precursor_app} also reports the individual stress components.

\paragraph{Prompt Variants Do Not Form Sharply Separated Regimes}
We cluster Retail prompt variants using dynamics derived from the world-model error trajectories defined by Equation~\eqref{eq:world_model_error}. The best silhouette score is $\mathbf{0.555}$ at two clusters, while the interpretable six-cluster partition is weaker. Figure~\ref{fig:universality_app} reports cluster quality and separation.

\paragraph{Interventions Depend on the Task}
We compare verification, exploration, memory, and planning interventions across ALFWorld and $\tau$-bench Retail using the same world-state measurements defined in Section~\ref{sec:preliminaries}. The highest mean reward is $\mathbf{0.433}$ in ALFWorld and $\mathbf{0.500}$ in Retail, but the corresponding intervention settings differ. Figure~\ref{fig:regimes_app} reports the full comparison.

\section{Conclusion}

We presented a twenty-two-experiment analysis of finite world-model SOC in long-horizon LLM agents. Our results show structured critical dynamics: increasing stress shifts agents from locally contained state errors toward persistent, dependency-mediated propagation, with avalanches constrained by finite horizons and task structure. However, the evidence does not support a universal or sharply defined critical regime: natural stress has limited predictive power, simple persistence explains part of the avalanche structure, and effective interventions vary across substrates. 

%
%

\bibliographystyle{ACM-Reference-Format}
\bibliography{references}

\clearpage
\appendix

\section{World-State Definitions}
\label{app:operationalization}

Section~\ref{ass:frozen} treats the agent-implied state $z_{i,t}$, gold state $z_{i,t}^\star$, and state distance $d_s$ abstractly. Here we specify how these quantities are constructed for each benchmark substrate. When the benchmark provides the required information, we represent the measured world state using seven interpretable components:
\begin{equation}
z_{i,t}=(q_{i,t},b_{i,t},c_{i,t},u_{i,t},r_{i,t},m_{i,t},p_{i,t}),
\end{equation}
where $q$ denotes task progress, $b$ the agent's current belief about the environment, $c$ constraint satisfaction, $u$ unresolved uncertainty, $r$ accumulated execution or tool risk, $m$ relevant interaction or retrieval memory, and $p$ the current plan or intended next subgoal. These components are not required to have the same representation across substrates; each is defined using benchmark-observable information appropriate to that environment.

Table~\ref{tab:state_components} gives the operational meaning of each component and the source used to construct the corresponding gold state $z_{i,t}^\star$. Components that are not observable or not meaningful for a substrate are excluded from its state distance rather than estimated from unavailable information. Count-valued components are normalized by the task horizon or a fixed component maximum, while set- or structure-valued components use normalized benchmark-specific distances. All component definitions and normalization rules are fixed before outcome analysis, as required by Assumption~\ref{ass:frozen}.

\begin{table*}[!ht]
\centering
\scriptsize
\resizebox{\textwidth}{!}{%
\begin{tabular}{lllllllll}
\toprule
Substrate & $q$ & $b$ & $c$ & $u$ & $r$ & $m$ & $p$ & Gold source \\
\midrule
StatefulPuzzle & subgoals & attribute table & violations & ambiguity & none & corrupted buffer & next subgoal & simulator log \\
Retail & subtasks & intent/db belief & policy violations & hedging & env errors & tool history & next action & $\tau$-bench gold \\
Airline & subtasks & booking belief & policy violations & ambiguity & tool errors & call history & next action & $\tau$-bench gold \\
GAIA & steps & evidence state & none & search uncertainty & none & retrieved docs & next reasoning step & GAIA answer \\
ALFWorld & subgoals & room/inventory & none & object uncertainty & none & observations & next subgoal & game log \\
HotpotQA & hops & answer hypothesis & none & retrieval uncertainty & none & passages & next retrieval/reasoning step & supporting facts \\
\bottomrule
\end{tabular}
}
\caption{\textbf{World-state definitions.} Each substrate instantiates the agent-implied state with benchmark-observable components and an independently constructed gold state. Unavailable components are omitted from the corresponding state-distance calculation.}
\label{tab:state_components}
\end{table*}

For each substrate $s$, the distance $d_s(z_{i,t},z_{i,t}^\star)$ therefore compares only the components defined for that substrate. This produces the world-model error $e_{i,t}$ and fidelity $F_{i,t}$ used throughout the main analysis.

\section{Additional Verification}\label{app:verification}

\paragraph{Independent-error baseline.}
We compare the observed collapse rate with an independent per-step error baseline. If the per-step error probability is $\varepsilon_0$, the probability of at least one error over horizon $H$ is $C_{\mathrm{ind}}(H)=1-(1-\varepsilon_0)^H$. Using $C_{\mathrm{ind}}=0.133$ and $H=64$ gives $\varepsilon_0\approx0.0022$.

\paragraph{Dependency-depth crossover.}
Table~\ref{tab:d_transition_app} provides additional statistics for the dependency-depth experiment in Figure~\ref{fig:chaos}. The mean $\Delta$AIC changes from negative at depth one to positive for all tested depths $D\geq2$, while the fraction of trajectory pairs favoring the power-law fit increases from $0.133$ to at least $0.667$. Saturation statistics are reported separately because divergence remains bounded.

\begin{table*}[!ht]
\centering
\small
\begin{tabular}{rrrrr}
\toprule
Depth & Power beats exponential & $\Delta$AIC exp-power & Steps to saturation & Mean divergence \\
\midrule
1 & 0.133 & -2.754 & 11.9 & 0.120 \\
2 & 0.667 & 0.648 & 10.5 & 1.895 \\
4 & 0.700 & 0.558 & 7.8 & 1.947 \\
6 & 0.667 & 0.136 & 6.9 & 1.878 \\
8 & 0.700 & 0.533 & 10.3 & 1.910 \\
\bottomrule
\end{tabular}
\caption{\textbf{Additional dependency-depth statistics.} Model-fit preference, saturation time, and mean divergence across dependency depths in StatefulPuzzle.}
\label{tab:d_transition_app}
\end{table*}

\section{Complete Evidence Accounting and Null Models}\label{app:accounting}
Table~\ref{tab:full_scoring} gives the complete evidence accounting used by the paper. It expands Table~\ref{tab:evidence_map} with headline metrics and interpretation notes.

\begin{table*}[!ht]
\centering
\scriptsize
\renewcommand{\arraystretch}{1.15}
\setlength{\tabcolsep}{5pt}
\resizebox{\textwidth}{!}{%
\begin{tabular}{@{}p{.18\textwidth}p{.32\textwidth}p{.16\textwidth}p{.28\textwidth}@{}}
\toprule
\textbf{Diagnostic} & \textbf{Main result} & \textbf{Evidence} & \textbf{Takeaway} \\
\midrule
Avalanche tails
& Truncated power law fits better than a pure power law
& Partial
& Error cascades are heavy-tailed but bounded \\

Belief-state persistence
& $P_{\mathrm{esc}}(\mathrm{wrong})=0.97$ vs.\ $P_{\mathrm{esc}}(\mathrm{success})=0.82$
& Alternative
& Incorrect states are shallow rather than strongly trapping \\

Local--global gap
& $\Delta^{LG}=0.857$ on GAIA and $0.62$--$0.68$ on Airline
& Supporting
& Locally valid actions can coexist with incorrect global state \\

Task-preserving changes
& KS distances remain between $0.03$ and $0.13$
& Supporting
& Collapse statistics remain stable under wording and format changes \\

Natural precursor signal
& AUROC $=0.616$
& Partial
& Naturally observed precursors provide limited predictive value \\

Temporal dependence
& $\alpha_e\in[1.38,1.52]$
& Supporting
& Errors show persistent temporal dependence \\

Dependency-depth transition
& Mean $\Delta$AIC changes sign at depth two
& Supporting
& Error propagation changes regime while remaining bounded \\

Graph-conditioned geometry
& $D_f=0.835$--$0.904$ on HotpotQA; $0.639$--$1.50$ on ALFWorld
& Supporting
& Error geometry depends on task structure \\

Prompt-regime structure
& $k=6$ silhouette score $=0.321$
& Partial
& Prompt variants form weakly separated rather than discrete regimes \\

Finite-size scaling
& $A_{\max}$ increases from $7$ to $490$
& Supporting
& Larger horizons support larger error cascades \\

Capability boundary
& Valid trajectories exist through $L=64$ but not for $L\ge96$
& Capacity-limited
& Dynamics cannot be identified beyond the valid-generation range \\

Intervention regime
& Best reward differs across Retail and ALFWorld
& Boundary
& No single intervention setting is optimal across substrates \\
\bottomrule
\end{tabular}
}

\caption{\textbf{Evidence summary across diagnostics.} Each row reports the main quantitative result, its evidence category, and the corresponding interpretation.}
\label{tab:full_scoring}

\end{table*}

\paragraph{Avalanche nulls.}
Retail avalanches differ from independent per-step errors. A matched Bernoulli process reproduces the mean error rate but underestimates clustering and large bursts, while a two-state Markov model captures more of the observed persistence \citep{feller1968introduction}. This comparison distinguishes independent errors from temporally correlated error cascades.

\paragraph{Heavy-tail model selection.}
We fit Retail avalanche sizes using finite-sample power-law procedures \citep{clauset2009power}. Lognormal and truncated power-law models fit better than a pure power law, indicating heavy but bounded tails rather than an unrestricted power law.

\paragraph{Stress-prediction nulls.}
For Retail, we predict later reward errors using stress measured earlier in the trajectory. Stress performs above chance, but task difficulty provides similar predictive power and adding stress gives only a small improvement. Natural stress is therefore a modest predictor of later failure, distinct from the controlled stress intervention in StatefulPuzzle.

\paragraph{DFA diagnosis.}
Short trajectories can produce unreliable DFA estimates. We therefore use DFA only at longer horizons and rely on spectral exponents across all horizons \citep{peng1994dfa}.

\paragraph{Multiple testing.}
Confirmatory tests use Benjamini--Hochberg false-discovery control \citep{benjamini1995fdr}. The main finite-size and long-horizon spectral results remain significant after correction.

\section{Additional Boundary and Direction Results}\label{app:moved_results}
This appendix reports additional results that clarify where the main dynamical patterns hold and where they become weaker or change direction.

\subsection{Shallow-Basin Direction}\label{app:part2_moved}
We test whether incorrect beliefs trap agents in ALFWorld. Agents leave incorrect states quickly, but also leave successful states frequently. The result therefore supports shallow and recoverable belief states rather than strong trapping.

\subsection{Retail Stress Precursor}\label{app:part5_moved}
We test whether early Retail stress predicts later reward errors. The signal is above chance but modest, and task difficulty performs similarly. Figure~\ref{fig:stress_precursor_app} compares stress with the difficulty baseline.

\subsection{Prompt-Regime Clustering}\label{app:part9_moved}
We test whether prompt variants form distinct dynamical regimes. The variants can be grouped into interpretable clusters, but the overall structure is continuous rather than clearly separated. Figure~\ref{fig:universality_app} shows the resulting clustering.

\subsection{Task-Dependent Intervention Regimes}\label{app:part11_moved}
The most effective intervention differs across tasks. Retail favors no intervention or stronger verification, whereas ALFWorld favors greater exploration or memory use. Figure~\ref{fig:regimes_app} compares these intervention settings.

\section{Discussion}\label{sec:discussion}

Our results suggest several additions to long-horizon agent evaluation. Benchmarks should track the world state implied by an agent, not only final reward or action validity. They should also measure local--global gaps and analyze error sequences over time. These measurements should be repeated across horizons, dependency depths, and task structures because error behavior changes with the interaction setting.

This goes beyond showing that errors accumulate. We ask how they accumulate: whether errors are independent or correlated, whether they form cascades, how earlier errors affect later states, and how these patterns change with task structure and horizon. The results show temporally correlated errors, bounded cascades, local--global gaps, and changes in propagation with dependency depth.

The three dynamical perspectives organize these observations. Self-organized criticality motivates tests of error cascades, temporal dependence, and finite-size scaling. Weak chaos describes how small state differences propagate with dependency depth while remaining bounded. Metastable belief dynamics describes how agents remain in or move between correct and incorrect belief states. Together, these perspectives provide a practical framework for studying how world models change during long-horizon interactions.

\end{document}